# Image Prior and Posterior Conditional Probability Representation for Efficient Damage Assessment

Jie Wei[1], Weicong Feng[1], Erik Blasch[2], Erika Ardiles-Cruz[2], Haibin Ling[3]
[1] City College of New York, [2] Air Force Research Lab, [3] Stony Brook Univ.

## Abstract

It is important to quantify Damage Assessment (DA) for Human Assistance and Disaster Response (HADR) applications. In this paper, to achieve efficient and scalable DA in HADR, an image prior and posterior conditional probability (IP2CP) is developed as an effective computational imaging representation. Equipped with the IP2CP representation, the matching pre- and post-disaster images are effectively encoded into one image that is then processed using deep learning approaches to determine the damage levels. Two scenarios of crucial importance for the practical use of DA in HADR applications are examined: pixel-wise semantic segmentation and patch-based contrastive learning-based global damage classification. Results achieved by IP2CP in both scenarios demonstrate promising performances, showing that our IP2CP-based methods within the deep learning framework can effectively achieve data and computational efficiency, which is of utmost importance for the DA in HADR applications.

## 1  Introduction

Effective Human Assistance and Disaster Response (HADR) [9] requires accurate damage assessment (DA). HADR DA is challenging because it is a catastrophic event, typically without prior exemplars from which resolved damage levels facilitate effective responses. Essentially, the HADR situation resembles an infrastructure of critical importance requiring substantial region monitoring and surveillance, robust operating conditions assessment, and knowledge of highly dangerous and contested scenarios. In all these situations, manual evaluations by human experts for prompt and real-time damage level assessments for the Object of Interest (OOI) [12] are nearly impossible and unsalable: The manual labeling process is extremely labor-intensive and time-consuming as each area of the OOI must be patiently studied to decide the correct labeling category wherein lots of expertise on the object types and damage styles are needed; which is even more grave for major disasters/actions over a wide area that hampers the HADR or DA efforts. Cutting-edge hardware, such as GPU, high-performance computing (HPC), and software, such as Artificial Intelligence and Machine learning (AI/ML), especially the deep learning (DL) [7] approaches should be called upon to assist users for OOI DA[14, 13].

Using the XView2 data (https://xview2.org), a large set of satellite overhead images was annotated. Recent results have been generated from overhead imagery challenges from winning teams [1], and new AI/ML methods continue to be explored [14]. Such advances should offer data and computational efficiency to render the work for practical utility. The methods should be both data and computationally efficient because 1) The available data in HADR applications are generally small which is significantly less than enough to adequately train a DL from scratch, and 2) The HADR personnel are not expected to have on-site access to immense computing power in their mission.

To design effective data and computing efficient HADR and DA approaches of practical utility, an Image Prior and Posterior Conditional probability (IP2CP) encoding scheme is developed to effectively encode OOI, e.g., buildings in HDAR or interested targets/regions in DA, before and after the disasters in HADR or actions in DA, respectively, to be used for damage level assessments. IP2CP is applied to two different scenarios: 1) pixel-wise semantic segmentation where the damage levels are densely classified for each pixel of the OOI, and 2) patch-based damage level classification (PDLC) for a small image patch centered around the OOI where a single damage level is assigned for the entire patch. Both scenarios are of crucial importance for HADR efforts: the pixel-wise segmentation filters out the most urgent regions from a wide span of regions; whereas the patch-level classification focuses on one crucial OOI, e.g., a hospital or school area, which may be of central importance to the mission, to make a judgmental call for a timely response.

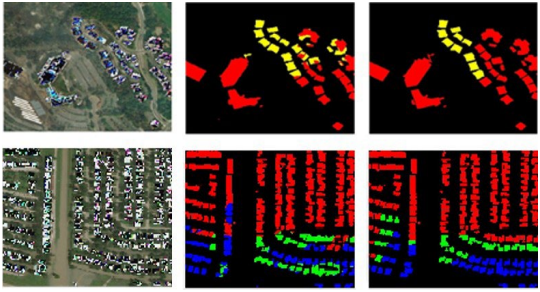

Figure 1: Two typical sample results of pixel-wise segmentation using IP2CP. Left column: IP2CP image, Middle column: pixel-wise labels predicted by IP2CP semantic segmentation; Right column: Ground truths (color code scheme: red: no damage, green: minor damage, blue: major damage, yellow: total damage). The $F_1$ scores delivered by the IP2CP-based segmentation approach for the top and bottom rows are 0.89, 0.64.

## 2   PROBABILISTIC IMAGING REPRESENTATION: IP2CP

To achieve effective damage level assessment, the images of the pre- and post-disaster/action for the OOI are normally available and well registered and matched. A contrastive Learning (CL)-based classifier learns the representation from pre- and post-disaster images using a Siamese network; CL is an effective method to attach contents of similar identities while pushing away those of dissimilar identities using Eq. (1) to define the loss function Loss $L_{CL}$ of a CL embedding network [5

$$\boldsymbol{L}_{CL} = \mathbf{1}\left[y_i = y_j\right] \|\theta\left(x_i - x_j\right)\|_2^2 + \mathbf{1}\left[y_i \neq y_j\right] \max\left(0, \epsilon - \|\theta\left(x_i - x_j\right)\|_2^2\right) \quad (1)$$

where $1[\cdot]$ is the indicator function, $x_i$ and $x_j$ are the two samples with corresponding labels $y_i$ and $y_j$ that need to be contrasted, $\theta$ is the embedding deep net, a 4-layer convolutional neural network (CNN) in our case, and $\epsilon$ is a hyperparameter dictating the lower bound distance between samples of dissimilar classes. For given training datasets, similar samples are generally produced by resorting to data augmentation, e.g., conducting image rotation, flipping, and color jittering, to create more instances of the same label ($y_i$'s); thus producing adequate training/testing data for effective CL training. The embedding net $\theta$ is trained by minimizing Eq. (1) to effectively align objects of the same labels closer while separating those of different labels.

CL is used in a broad spectrum of applications recently [6]. However, one major challenge of the CL framework in the XView2 dataset, and actually for DA efforts in HADR applications, is that: in the large scene covered by the imageries, the OOI only covers a small portion of the entire scene, e.g., more than 95% of the pixels on average in the XView2 dataset belong to the background that is of no relevance to the damage level assessment task. In consequence, it is highly questionable that the representations learned via CL, i.e., by minimizing the loss function dictated by Eq. (1), are those from the OOI, but most likely are about the background, which accounts for about 95% of the scene and is of little concern to HADR. Therefore, simply using the powerful CL may not achieve desirable results in HADR/DA applications. One major recent breakthrough in DL is the induction of the attention mechanism [11], where the valuable processing is focused more on those pixels of salience to the OOI, which has since revolutionized Natural Language Processing and Computer Vision with transformers [8]. Unless we can find a means to put the attention of the CL to the OOI, the prowess of CL cannot be fully exploited, which was evidenced by low classification performances as shown in Table 1.

Given the foregoing observations, IP2CP was developed to generate image $Z_c$ from pre-image $X_c$ and post-image $Y_c$, viewed as random variables with the following set of two conditional probability formulas:

$$P(Z_0 \mid X_0, Y_0) = Y_0 \quad (2)$$

$$P(Z_1 \mid X_1, Y_1) = \text{Norm}(Y_1 - X_1) \quad (3)$$

where $X_c$, and $Y_c$ are the pre- and post-image with values in the range [0,1], subscript c is the image indicator variable: 0 for background, 1 for OOI, $Z_c$ is the new random variable of the same range as X and Y, and Norm(w) is defined below:

$$\text{Norm}(w) = \frac{w}{\max(w) - \min(w)} \quad (4)$$

which brings the random variable of pixel (w) to [0,1]. Eqs. (2,3,4) essentially generate a new image Z whose background comes from the post image, the OOI is the normed version of the difference in the OOI region. By assessing the difference in the range [0,1], the pixel-wise label variations in the OOI are stressed. On the leftmost panel of Fig. 1, the IP2CP is shown as an illustration, where it can be observed that the OOI regions, i.e., the buildings, are of more outstanding color variations differing from the background. There are several benefits of using this new imaging representation Z instead of the original X or Y: 1) Data and computing efficiency: instead of two color images, now there is only one to be processed, reducing the data and computing loads. In contrast, the CL is extremely RAM and GPU-hungry to process big sets of pre- and post-disaster images. 2) Far more choices of DL methods: with merely three channels (red, green, blue), methods have more choices than original DL or CL facilitating 6-channel deep nets for data classification, in both semantic segmentation and image/object classification. 3) Emphasis on the OOI: due to Eqs. (2, 3), the OOI regions have larger values and variances and thus could be given more importance in data training. 4) Simulation of human annotation: in manual labeling by human experts, the pixel-wise difference in OOI plays an important role in making the damage level assessment. 5) Contextual information: Eq.(2) states that the post images provide the most relevant contextual information to the OOI assessment, as the situations after natural disasters or first-responder actions carry more weighty information on the OOI than those before the disasters or actions. However, using the pre-disaster image or removing the background resulted in far worse classifications per our tests. The post-disaster/action information should thus be employed as dictated by Eq. (2) to provide valuable contextual information.

In Eq.(3), we use a simple differentiation operator to obtain the best contrast during the disasters between the pre- and post-disaster images for HADR, as the XView2 datasets are perfectly matched by satellite longitudes and latitudes. In more general cases where such perfect matches are unavailable, a matching procedure using conventional computer vision techniques such as SIFT features combined with the RANSAC process [10] and DL-based ones [15] can be first conducted to obtain IP2CP, our initial tests of using approximate visual matching instead of given perfect matches from XView2 suggested similar performances. We also tried using CL and attention layers to learn such attention-like representations [11], but in vain: far worse performances are consistently observed despite a long training procedure induced therein.

## 3 IP2CP applications

In this section, the new IP2CP representation will be applied to two different situations: the first one is for global semantic segmentation as originally done for HADR: identify the damage levels for the entire image for every OOI. The second is for patch-based classification: the OOIs, e.g., hospitals or schools, as the localized damage assessment for these important regions should be evaluated first with high accuracy.

### 3.1 IP2CP in semantic segmentation

For images covering a wide expanse, as shown in the leftmost panel of Fig. 1, the semantic segmentation procedure is needed to generate dense pixel-wise damage classification. One of the challenges in deep learning deployment is to determine the best method. Instead of choosing a single best method, which is hard to determine, robust approaches endeavor to seek out a fusion of classifiers [5]. For our method, the U-Net is combined with transfer learning, which has been exceedingly powerful in image segmentation. Besides CL, it is yet another method to achieve data and computation-efficient classification. Using Python's segmentation models or Matlab©'s segnetLayers methods on the XView2 dataset, the newest DeepLabv3+ network [2] yields reasonably competitive results. The U-Net based on which can be initialized by ImageNet-trained Resnet or Mobilenet, valuable knowledge learned from ImageNet can be readily transferred to the disaster overhead images in the XView2 dataset. In this combination of U-Net, DeepLab v3+, and DL architecture, depth separable convolutions are used in the Atrous spatial pyramid pooling and decoding subnets. To further augment training, various scales and geometric transforms such as rotations, flipping, and shearing are employed on the fly during training.

Table 1: Pixel-wise semantic segmentation performances

| Alg. | RM1 | RM2 | Post | CL/TR | IP2CP |
|---|---|---|---|---|---|
| $F_1$ (%) | 49.7 | 61.4 | 66.4 | 67.4 | 69.7 |
| Size (MB) | 441 | 228 | 9.7 | 40 | 9.7 |

Equipped with IP2CP, which can be treated as a set of normal color images with OOI stressed, by taking advantage of transfer learning: using pre-trained deep nets such as ResNet for feature generation and DeepLabV3+ for U-net-based pixel-wise segmentation, a dense pixel-wise semantic segmentation was performed to label all pixel in the overhead images. In Table 1, the semantic segmentations performances delivered by the new IP2CP-based method and two reproduced No. 1 and No. 5 winning methods (RM1 and RM2) in XView2 challenge [1] and two our own methods: the transfer learning based on post-disaster data only (Post) and CL combined with transfer learning (CL/TR) are reported. The IP2CP captured more OOI information than the Post and CL/TR, thus delivering the best classification performance (0.697). Furthermore, as the IP2CP is only one color image, the deep nets are considerably smaller than the CL-based method. RM1 and RM2 are the results of reproducing the winning methods in the original XView2 challenge. The considerable performance drops are most likely due to the reduced resolution of the XView2 datasets: we reduced them from 1024×1024 to 512×512 for data efficiency: without which we cannot conduct our tests on laptop computers with ordinary GPU and RAM size. Hence, this IP2CP-based semantic segmentation method is more data and computing efficient. In Fig. 1, two typical sets of segmentation results, with $F_1$ measures at 0.89 and 0.64, respectively, are illustrated.

## 3.2 Patch-based Damage Level Classification

The pixel-wise dense damage classification reported in the preceding subsection is used for images covering large areas. In HADR applications when of the utmost interest is the DA for a specific OOI in a small region, e.g., in HADR for a specific building one would like to know if it is damaged and needs urgent help. In these important scenarios, according to HADR personnel [3], the dense pixel-wise damage level classification, which is generally of relatively poor performance (with $F_1$ <0.7 with all state-of-the-art methods), is no longer needed; conversely, only a binary classification label is needed for the entire small patch where the OOI is located, whereof better classification results are needed and can indeed be expected if an appropriate approach is employed.

In the XView2 dataset, overall there are five different labels, which is one main reason why outstanding segmentation classification performances are elusive is due to the considerable unbalanced nature of these labels: for regions with damage covering only 5% of total area, there are about 90% are "no damage" (label 1), only <2% with "total damage". Various image augmentation methods were explored, e.g., GAN [4], different class importance, trying to mitigate the data imbalance problem, but with little success. The best $F_1$ score by IP2CP is still slightly less than 0.70. In the patch-based scenario, the situation can be significantly improved: 1) Instead of separating damage levels from "minor" up to "total", a binary label "no damage" and "with damage" will suffice for many practical HADR scenarios, especially for OOI of crucial importance [3], then the trouble caused by the extremely small number of "major" and "total" damage levels, which gravely hampered the pixel-wise DA efforts, can be effectively avoided. 2) When generating the small patches of "no damage" and "with damage", a statistics collection procedure is conducted by varying the patch size (128×128, 64×64, etc.) and OOI size of different types, then fixing the two thresholds ($\delta_1$, $\delta_2$) for "no damage" and "with damage" types.

For a patch p, its class label is "no damage" or "with damage" if the maximal OOI of p has its size exceeds $\delta_1$ and $\delta_2$, respectively, and the OOI regions belonging to the other class are erased to ensure the current patch p only has the winning label to be identified—without which the OOI of the other class in p will essentially confuse the training process. The method produces patches with these two types of roughly similar size to facilitate more effective future classification [6]. From the XView2 datasets, after intensive offline processing, it is found that by using 64×64 patches and setting ($\delta_1$, $\delta_2$) to (0.12, 0.04), the training patch number for the two labels are 21k and 20k; and for the testing set there are 7.1k vs. 6.8k for these two classes, hence no serious class unbalances exist anymore, thus facilitating balanced training and testing processes.

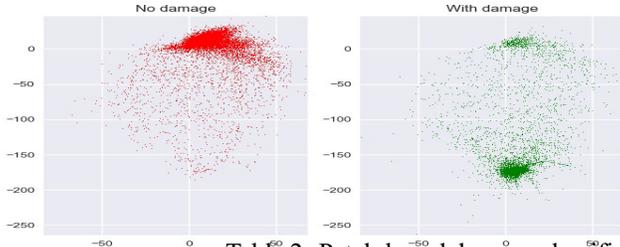

Figure 2: Contrastive learning 2D embedding for IP2CP patches labeled "No damage" (red points) and "With damage" (green points), which are mostly pushed away by a wide margin and can thus be easily separated by a simple classifier, such as a support vector machine in Eq. (5).

Table 2: Patch-based damage classification performances

| Alg. | $F_1$(%) | Net Size (KB) |
|---|---|---|
| ViT | 95.0 | 343,261 |
| VGG19 | 94.2 | 558,326 |
| ResNet152 | 93.0 | 233,503 |
| Inception V2 | 90.5 | 100,810 |
| googleNet | 90.3 | 22,585 |
| IP2CP-CL | 95.9 | 8,566 |

The IP2CP patch is now a 64×64 color image/tile with binary identities: with or without damage. Now that the patches in the same class are similar, whereas those belonging to a different class are dissimilar, CL as described in Sec. 2 can thus be effectively applied to embed patches of the same class in a low-dimensional space of a small distance, meanwhile pushing away those of different classes. In our simple Siamese network, the CNN consists of 3 convolution layers followed by a fully-connected layer to embed each 64×64×3 color image into the 2D space by minimizing the twin contrastive loss dictated by Eq. (1), the hyper-parameter $\epsilon$ is set at 2. After training the CL (50 epochs are adequate from our experiments), we use the trained CNN $\theta$ to embed all training sample data x's to a 2D space, and then fit a simple Support Vector Machine as the resultant classifier from the training process. With the given training datasets and the CL net, the predicted label $y_{pred}(x_i)$ for each patch $x_i$ encoded by CNN $\theta$ of the testing dataset can be obtained by the following equation:

$$y_{pred}(x_i) = \text{SVM}(\theta(x_i)) \qquad (5)$$

The resultant $F_1$ score we achieved using this approach for our testing dataset is 0.959 with the corresponding confusion matrix ($\begin{smallmatrix} 0.98 & 0.02 \\ 0.06 & 0.94 \end{smallmatrix}$).

As depicted in Fig. 2, the two classes in the test dataset are separated well in the 2-D embedding space reached by CL. We also tried out different embedding space dimensionality other than 2, e.g., from 3 to 10, of which the performances are all slightly worse ($F_1$ of 0.85 to 0.93) than the simple 2-D space. Thanks to IP2CP representation, this patch-based damage level classification is similar to the original ImageNet, transfer learning from readily available deep nets can be employed. CL is especially powerful as it not only delivered the top $F_1$ score but also with a net size significantly smaller: 9.5 MB vs. 343 MB, as reported in Table 2, where 15+ different nets are tried, and only the top 5 with $F_1 > 0.9$ are reported.

IP2CP combined with CL, therefore, achieved exceptionally valuable data and computing efficiency in this patch-based damage assessment task, as the IP2CP directed the data and computing attention to the desired regions to exploit the immense discrimination power of CL for data classification.

## 4 Conclusion

This paper presents a new representation using image prior and posterior conditional probability to effectively stress the changes in the object of interest. With the representation, outstanding pixel-wise damage classification on semantic segmentation and patch-based damage classification performances are consistently observed. This representation combined with various deep learning methods such as deep U-net and contrastive learning, together with conventional machine learning methods such as support vector machines, has yielded encouraging segmentation and classification performances with a considerably small size, which is of importance to practical use cases in HADR.